\pdfoutput=1

\documentclass[11pt]{article}

\usepackage[]{EMNLP2022}

\usepackage{times}
\usepackage{latexsym}

\usepackage[T1]{fontenc}

\usepackage[utf8]{inputenc}

\usepackage{microtype}

\usepackage{inconsolata}

\usepackage{graphicx}
\usepackage{amsmath, bm}
\usepackage{amssymb}
\usepackage{booktabs}
\usepackage{float}
\usepackage{paralist}
\usepackage{enumitem}

\usepackage[capitalize]{cleveref}
\crefname{section}{Sec.}{Secs.}
\Crefname{section}{Section}{Sections}
\Crefname{table}{Table}{Tables}
\crefname{table}{Tab.}{Tabs.}

\newcommand{\R}{\mathbb{R}}
\usepackage{xcolor,colortbl}

\title{MAGMA -- Multimodal Augmentation of Generative Models through Adapter-based Finetuning}

\author{Constantin Eichenberg\thanks{ \ \ Equal contribution} \and Sidney Black\footnotemark[1] \and Samuel Weinbach \\
Aleph Alpha \\
\texttt{\{constantin.eichenberg, samuel.weinbach\}@aleph-alpha.com, sdtblck@gmail.com} \AND 
Letitia Parcalabescu \and Anette Frank \\
Heidelberg University \\
\texttt{\{parcalabescu, frank\}@cl.uni-heidelberg.de}}

\begin{document}
\maketitle
\begin{abstract}
Large-scale  pretraining is  fast  becoming  the  norm  in Vision-Language (VL) modeling. However, prevailing VL approaches are limited by the requirement for labeled data and the use of complex multi-step pretraining objectives. We present $\mathrm{MAGMA}$ -- a simple method for augmenting generative language models with additional modalities using adapter-based finetuning. Building on \textit{Frozen} \cite{frozen}, we train a series of VL models that autoregressively generate text from arbitrary combinations of visual and textual input. The pretraining is entirely end-to-end using a single language modeling objective, simplifying optimization compared to previous approaches. Importantly, the language model weights remain unchanged during training, allowing for transfer of encyclopedic knowledge and in-context learning abilities from language pretraining. $\mathrm{MAGMA}$ outperforms Frozen on open-ended generative tasks, achieving state of the art results on the OKVQA benchmark and competitive results on a range of other popular VL benchmarks, while pretraining on $\sim 0.2 \%$ of the number of samples used to train SimVLM~\cite{simvlm}.
\end{abstract}

\section{Introduction}\label{sec:intro}

Self-supervised representation learning with transformer models \cite{vaswani2017attention} has become the dominant technique in Natural Language Processing in recent years, with encoder transformer models trained using a Masked Language Modeling (MLM) objective \cite{bert} excelling at Natural Language Understanding tasks, and autoregressive decoder models \cite{gpt, gpt2, gpt3} displaying impressive Natural Language Generation at increasingly large scales. Vision Language (VL) modeling -- the modeling of joint image-text representations for tasks such as image captioning or visual question answering (VQA) -- has followed suit, with the transformer encoder becoming the prevalent architecture in recent research. A popular approach among the latest state of the art VL models is to use a BERT-style encoder language model (LM) in combination with an object detection backbone such as Faster-RCNN \cite{faster-rcnn}.  This approach, while displaying impressive performance on challenging benchmarks, has a number of drawbacks (see Section~\ref{sec:related}), in particular not being able solve VL tasks in an open-ended, generative fashion.

\begin{figure}[t]
    \centering
    \includegraphics[width=\linewidth]{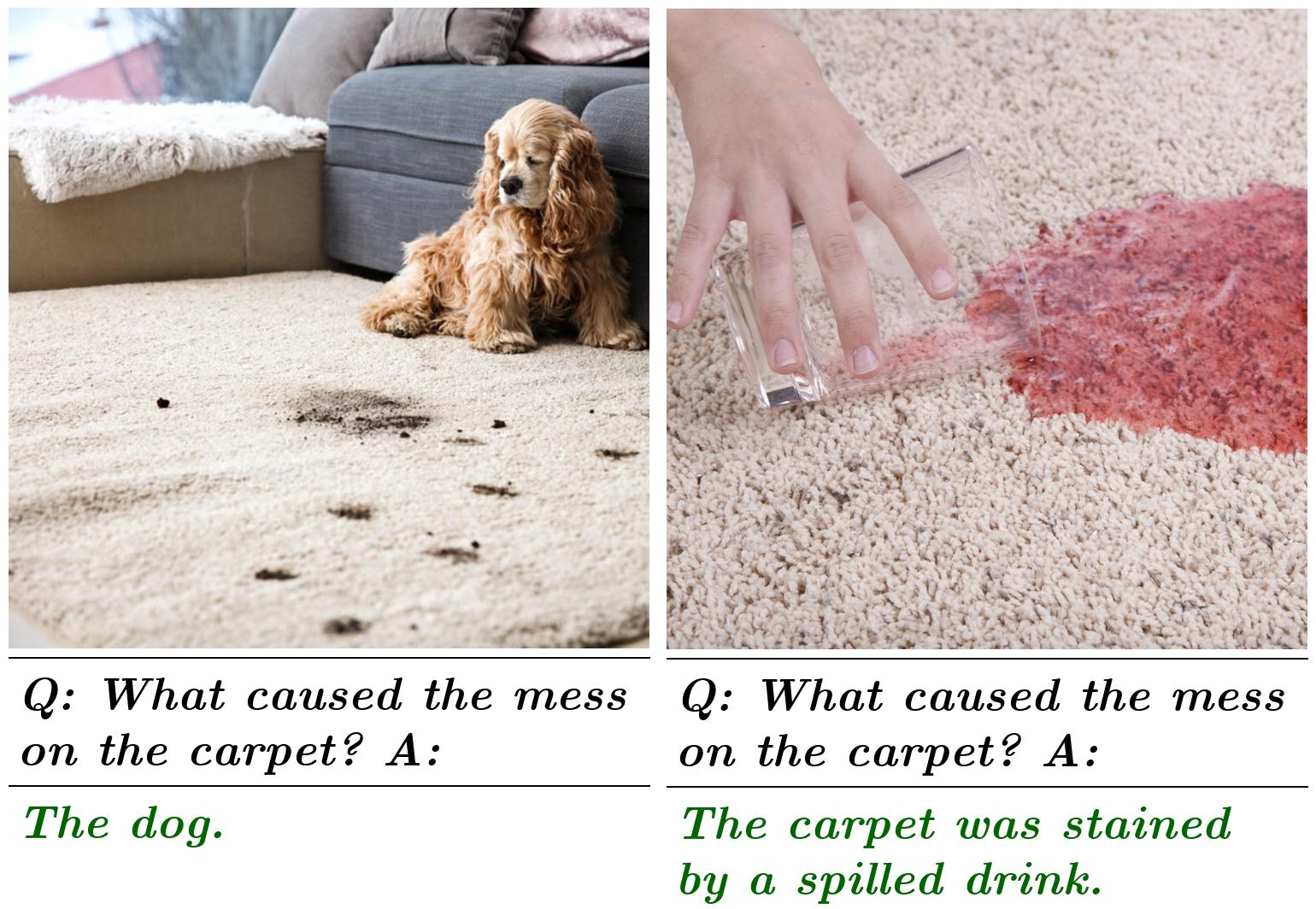}
    \caption{An example output produced by $\mathrm{MAGMA}$.
    For this and all following examples the input text is displayed in black, and the model's response in green.
    }
    \label{fig:figure_1}
\end{figure}

A recent line of work \cite{frozen, simvlm, mantis} explores VL modeling using autoregressive decoder models trained with a language modeling objective. \textit{SimVLM} \cite{simvlm} shows impressive performance, but requires prohibitively large amounts of pretraining data and the training of language and vision components in tandem. \textit{Frozen} \cite{frozen} shows that a pretrained autoregressive language model can, without any finetuning to the LM weights themselves, be harnessed to train a visual prefix which enables images to be used as its input. While the performance of \textit{Frozen} on VL benchmarks falls short compared to the state of the art, we feel the approach is promising due to its practicality, and the public availability of large, pretrained LMs such as GPT-J \cite{gptj}, PanGu-$\alpha$ \cite{pangualpha}, and GPT-Neo \cite{gptneo}.

Extending the \textit{Frozen} approach, in this paper we introduce a framework to combine existing unimodal language and unimodal vision models pretrained on large web datasets into a powerful multimodal model. Specifically, our contributions are:

\begin{enumerate}[label=\roman*), noitemsep,]
    \setlength\itemsep{0.1em} 
    \item We introduce $\mathrm{MAGMA}$: An autoregressive VL model that is able to generate text from an arbitrary combination of visual and textual input. Like \textit{Frozen}, we start from a fixed large LM and a visual encoder-prefix stack.~$\mathrm{MAGMA}$ differs from \textit{Frozen} by additionally augmenting the LM with adapter layers, and using CLIP's \cite{clip} visual component as encoder. Only training the adapters and visual components, the method is parameter efficient and naturally retains the LM's encyclopedic knowledge and \textit{in-context} learning abilities.
        
    \item Pretrained on a simple next token prediction objective,  $\mathrm{MAGMA}$ is competitive in several VL downstream tasks, significantly outperforming its predecessor,  \textit{Frozen}, while pretraining on $\sim$0.2 \% of the number of samples used for \textit{SimVLM}~\cite{simvlm}. In particular,  $\mathrm{MAGMA}$ achieves state of the art accuracy on the OKVQA benchmark, which we evaluate as a fully open-ended generative task.
    \item Our extensive ablations on the vision encoder and adapter components show i) that a pretrained CLIP ResNet encoder outperforms other visual backbones, ii) that an adapter-tuned model outperforms a visual prefix-only method, and iii) that different adapter configurations excel at different downstream tasks.
    \item We show that a carefully curated pretraining dataset -- including around $25$ million image-text pairs from a wide range of sources, including downstream task training data -- can dramatically increase downstream performance when compared to a noisier, web-scraped dataset (CC12M \citet{cc12m}).
\end{enumerate}

We only explore the VL domain in this work, but we expect the general method of a modality-specific prefix in combination with adapter layers and a frozen LM to apply equally well to other combinations of modalities, such as audio-text pairs.

With this publication, we open-source our code and release a trained model checkpoint.\footnote{\url{https://github.com/Aleph-Alpha/magma}}
\setlength{\belowcaptionskip}{-10pt}

\begin{figure*}
    \centering
    \includegraphics[width=\linewidth]{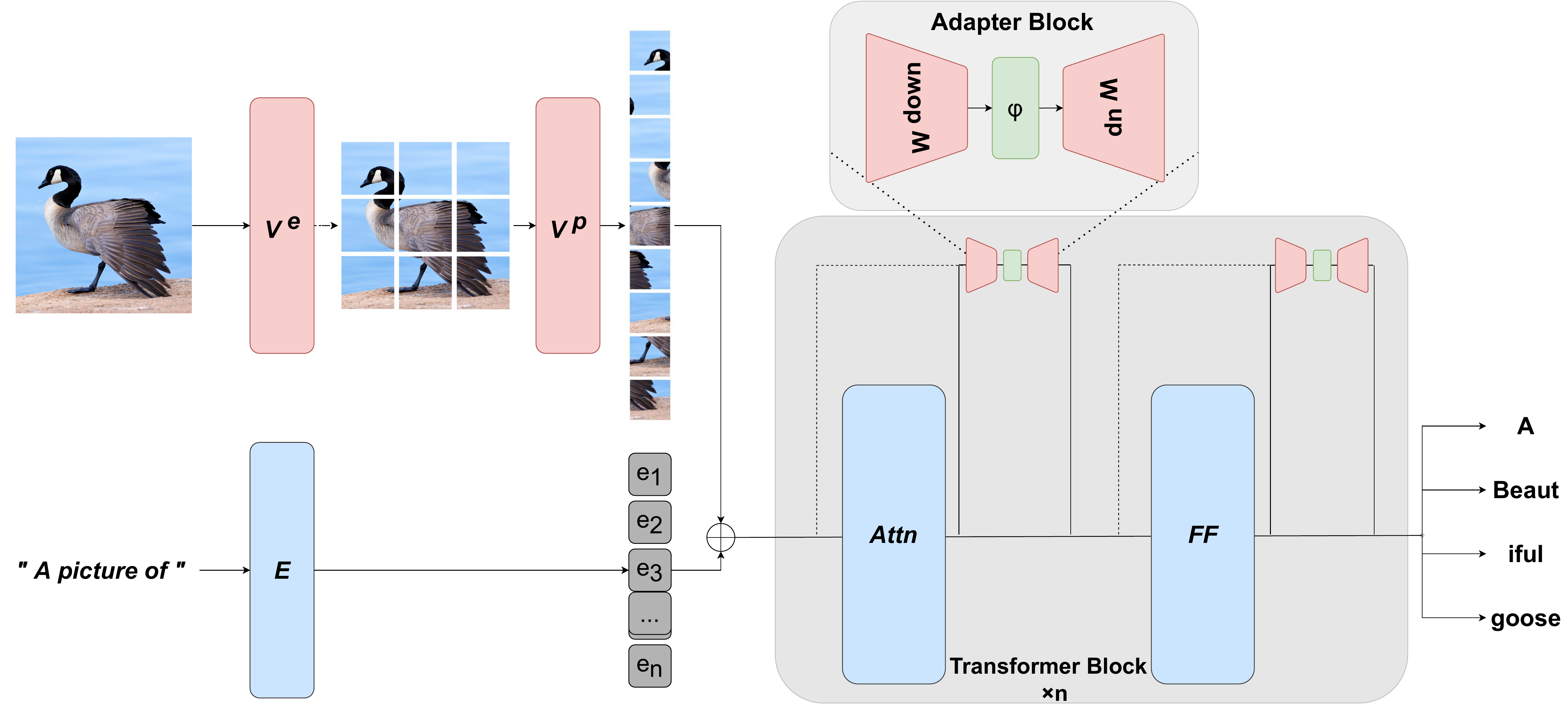}
    \caption{$\mathrm{MAGMA}$'s architecture. The layers in red are trained, and the layers in blue remain frozen.}
    \label{fig:model_schematics}
\end{figure*}

\section{Related Work}
\label{sec:related}

VL models of the past years \cite{vinvl, oscar, uniter, visualbert, vlbert, lxmert} harness a BERT-like encoder transformer as the language component, trained with a MLM objective -- where random words in the input are masked out, and the model is tasked with predicting them. Encoder VL models are often also pretrained with auxiliary objectives or custom cross-modal losses, such as the Masked Region Modeling, Image-Text Matching and Word-Region Alignment of UNITER \cite{uniter}, or the contrastive loss of OSCAR \cite{oscar}. Using auxiliary cross-modal loss functions and pretraining tasks complicates the pretraining procedure by requiring these losses to be properly balanced. Additionally, encoder models need extra task-specific finetuning for each task to perform effectively, limiting their accessibility. In comparison, autoregressive VL models like $\mathrm{MAGMA}$ are trained on a single, simple next token prediction objective, and can perform well on a wide range of tasks without further finetuning.

Two predecessors to our method are \textit{Frozen} \cite{frozen} and \textit{SimVLM} \cite{simvlm}, two autoregressive decoder models trained with a next token prediction language modeling objective. \textit{Frozen} affixes an NFResnet \cite{nfresnet} vision encoder to a pretrained autoregressive LM and, keeping the LM weights frozen, trains the vision encoder along with a \textit{visual prefix} that linearly maps the output of the vision encoder to the dimensionality of the LM's token embeddings. \textit{Frozen} shows that autoregressive VL models have the ability to adapt to examples \textit{in-context}, like their language only counterparts \cite{gpt3}, without performing any gradient updates. When shown multiple examples of a task in its context window in \textit{Few-Shot learning}, its performance on that task improves,
it appears to `learn' from the presented examples 
without task-specific finetuning. Our model has similar \textit{in-context} learning capabilities, but the addition of adapters and the different choice of visual backbone results in a model with improved performance when trained on a comparable dataset, see Section \ref{subsec:dataset}.

\textit{SimVLM} is similar to \textit{Frozen}, but pretrains the vision and language components in tandem using a prefix LM objective. \textit{SimVLM} consists of an encoder-decoder transformer with a combined ResNet \cite{resnet} and ViT \cite{vit} backbone as encoder, and a transformer decoder for language modeling. It extends the state of the art on a wide range of benchmarks, showing that a simple language modeling task can outperform MLM approaches. However, the joint pretraining requires prohibitively large uni- and multimodal datasets (1.8 Billion+ image-text pairs and $\sim$800GB of raw text), and long training times ($\sim$4 Billion image-text pairs and $\sim$130 Billion text tokens).
Aside from using orders of magnitude less data than \textit{SimVLM}, $\mathrm{MAGMA}$ allows for the full recovery of the underlying LM's performance by simply removing the adapter layers.

Our work
builds on recent advances in parameter efficient finetuning of LMs \cite{houlsby_adapters, prefixtuning, prompttuning, unified_adapters, lora}, specifically with adapter layers \cite{houlsby_adapters, unified_adapters, pfeiffer_adapters}, which are small modules inserted in between the elements of a transformer layer which are finetuned instead of the model weights as a form of parameter efficient fine-tuning.

For a visual backbone, it is common to use region features from a pretrained object detection model such as Faster-RCNN \cite{faster-rcnn}.
These are generally trained using expensive human labeled data on a bounded set of object classes, limiting the number of object types the resulting model can recognize.
On the other hand, contrastive models such as CLIP \cite{clip} and ALIGN \cite{align} present a more robust approach to learning visual features by learning joint representations between image-text pairs. They show strong performance on a wide variety of vision tasks as well as impressive generalization abilities that can provide powerful semantic guidance to image generation \cite{vqgan}. But since they were only trained to match image-text pairs, they cannot inherently be used for tasks that require text generation as output \cite{clip_vl}. 

However, \citet{clip_vl} show that the weights of contrastive language-image models contain useful semantic information for VL tasks. By replacing the conventional region-based backbone with CLIP's visual encoder in popular VL architectures, the authors achieve SOTA results across a wide variety of VL tasks without needing region-based features, motivating us to use CLIP's visual component as a vision encoder for $\mathrm{MAGMA}$. Notably, we confirm their finding that the ViT variant of CLIP underperforms on VL tasks when compared to the ResNet variant, particularly in tasks that require localization within an image.

\section{Method}
\label{sec:method}

Our general approach is an image conditioned variant of soft-prompting or prefix tuning \cite{prompttuning, autoprompt, softprompt} for language transformers and extends the \textit{Frozen} method ~\cite{frozen}. The core idea is to translate image features into language embeddings
carrying visual information which
can therefore be interpreted by the language transformer without need to retrain the latter from scratch.

\subsection{Architecture} \label{subsec:architecture}
The model can be broken down into four main components, see Figure \ref{fig:model_schematics}. First, images are fed into a \textit{Visual Encoder}, which processes the raw image input and outputs a sequence of feature vectors. Then an \textit{Image Prefix} module maps image features into a sequence of embedding vectors that are input to the third model component, an auto-regressive \textit{Language Model}. The fourth component is a series of \textit{Adapter} layers which are inserted into the transformer LM, and tuned during training. We discuss the four components in more detail below.

\paragraph{Visual Encoder -- $V^{e}$} The visual encoder is a network used to extract condensed semantic information about an image. In principle, the visual encoder could take the form of any deep vision network whose output can be mapped to a sequence of embedding vectors. For our ablations, we use the visual backbone of several variants of CLIP. We also train a model with an NFResnet encoder trained from scratch, which is analogous to the model presented in \textit{Frozen}, see §\ref{subsec:ablations}. The visual encoder output is then passed into the \textit{Image Prefix}.

\paragraph{Image Prefix -- $V^{p}$} Before the encoder output can be input to the LM, it needs to be translated into a sequence of $n$ $d_h$-dimensional vectors, where $d_h$ is the LM's hidden dimension. For the CLIP encoders, we extract the feature grid before the pooling layers, resulting in an $N \times N$ grid, where $N = 7, 7, 12$ for the ViT-B/32,  RN50x4 and RN50x16 variants of CLIP respectively. We then flatten the feature grid into a sequence of $N^2$ vectors, and linearly transform the vectors' channel dimension to $d_h$. For the NFResnet variant, we follow the procedure described in \textit{Frozen} by linearly transforming the output to  $d_h \cdot n$, where $n$ can be an arbitrary sequence length which we set to $2$. Finally, we apply dropout regularization to the output of the image prefix, followed by Layer Normalization. We also explored non-linear variants of prefix mappings, replacing the linear transformation with an MLP and a transformer encoder, but found no improvements. 

\paragraph{Language Model -- $E, T, H$} The language backbone of our architecture is initialized from a pretrained auto-regressive transformer LM similar to GPT \cite{gpt}. 

A text input $y$ is converted into a sequence of tokens $t_1,...,t_m$. Then a word embedding layer $E$ maps each token $t_k$ to a unique vector $e_k = E(t_k) \in \R^{d_h}$, obtaining a sequence of embeddings $e_1,...,e_m$ which are input to a transformer-decoder module $T$ with a causal attention mask. A language model head $H$ maps the final output embeddings of the transformer to logits over the token space which can be used in a cross-entropy loss function for a next-token-prediction training objective and to auto-regressively generate text during inference. Because any sequence of vectors $v_1,...,v_m \in \R^{d_h}$ can be used as input to the transformer, we can use images as input after mapping them through the encoder and the prefix as described above.

For the LM component, we use the open sourced weights of the 6 Billion parameter GPT-J \cite{gptj} LM. Since its architecture is largely similar to that described in \citet{gpt}, we will not cover it in this paper, but do note two key differences of GPT-J compared to the original GPT architecture. Firstly, GPT-J replaces learned positional embeddings with rotary positional embeddings \cite{rotary}, a form of relative positional embedding. As noted in \cite{frozen}, relative positional embeddings enable the transformer to generalize to inputs with more than one image, or a different image-text ordering compared to the training distribution, which is key to the VL model's ability to perform in-context learning with multiple image examples. Secondly, the attention layer and the feedforward layer are computed in parallel for decreased communication costs \cite{gptj}.

\paragraph{Adapters -- $\{A_{i} \}$} Adapters are a series of small modules placed in between elements of a transformer model~\cite{houlsby_adapters}, that can be finetuned instead of the model weights as a form of parameter efficient fine-tuning. We use the framework of~\citet{unified_adapters}, where the adapter layers take the form of a scaled residual bottleneck MLP:
\begin{align}
A_i(h) = h +  \lambda_i W^{up}_i\varphi \left (W^{down}_i h \right).
\end{align}
The matrices $W^{down} \in \R^{d_b \times d_h}$ and $W^{up} \in \R^{d_h \times d_b}$ with $d_b < d_h$ constitute the bottleneck, $\varphi$ is an activation function (in our case ReLU) and $\lambda_i$ is a scaling parameter that is either trained or set equal to $1$. We refer to the ratio $d_h / d_b$ as the \textbf{downsample factor} of the adapter.

Given a set of adapters $\{A_{i} \}$ and a transformer module $T$, we denote the adapted version of $T$ by $\Tilde{T}$, which means replacing the attention and/or feed-forward blocks $B_i$ of $T$ by their adapted version $\Tilde{B}_i$, either obtained from adding the adapters in parallel or sequentially:
\begin{align} \label{e:adapter}
\Tilde{B}_i \colon  h &\mapsto 
\begin{cases}
B_i(h) + A_i(h) \quad &\mathrm{(parallel)}  \\
B_i(h) + A_i(B_i(h)) \quad &\mathrm{(seq.)}
\end{cases}
\end{align}
We experiment with both parallel and sequential adapter variants, see Section \ref{subsubsec:adapter_ablations} for results.

\subsection{Training}
During training, the weights of the LM $E, T, H$ remain unchanged, whereas the weights of the image encoder $V^e$, image prefix $V^p$ and the adapters $\{A_i\}$ are optimized. The language model components are initialized with weights from the pretrained GPT-J model and the image encoder is initialized with pretrained CLIP weights except for the NFResnet ablation, where the image encoder is randomly initialized. The image prefix and adapters are always trained from scratch. In the following we denote the trainable parameters of a module by the subscript $\theta$. As described in~\ref{subsec:architecture}, a set of trainable adapters $\{A_{i,\theta} \}$ gives rise to the modified transformer module $\Tilde{T}_\theta$.

\textbf{The training objective is a captioning task}: given an image-caption pair $(x,y)$, we embed the image as $v_{1,\theta},...,v_{n,\theta} = V^p_\theta \circ V^e_\theta(x)$ and the text as $e_1,...,e_m = E(t_1),...,E(t_m)$, where $\{t_k\}$ is the tokenized caption $y$. Note that the image sequence length $n$ is fixed while the length of the caption $m$ is variable. The image embeddings are then prepended to the text embeddings and fed through the adapted transformer module. Denoting the embedding-to-logits function as $l_\theta = H \circ \Tilde{T}_\theta$, we then compute the loss 
\begin{align} \label{e:loss_fn}
L_\theta(x,y) = - \sum_{i=1}^m l_\theta(v_{1,\theta},...,v_{n,\theta},e_1,...,e_i),
\end{align}
where $l_\theta(v_{1,\theta},...,v_{n,\theta},e_1,...,e_i)$ is interpreted as next-token log-probability conditioned on the previous sequence elements
\begin{align}
&l_\theta(v_{1,\theta},...,v_{n,\theta},e_1,...,e_i) \notag \\
&= \log p_\theta(t_i \; | \; x, t_1,...,t_{i-1}).
\end{align}

For technical details regarding training, see \ref{app:appendix}.

\subsection{Dataset} \label{subsec:dataset}
For \textbf{pretraining} we use two different large scale datasets, one for the ablations and another one for our final model $\mathrm{MAGMA}_{base}$, respectively $\mathrm{MAGMA}_{long}$. 
For all \textbf{ablations} in \ref{subsec:ablations} we train on CC12M \cite{cc12m} for a total of around 3M samples ensuring comparability with \textit{Frozen}. Unfortunately, CC12M performs hypernyming, replacing people names with $\langle \textrm{PERSON} \rangle$. This causes downstream models to output $\langle \textrm{PERSON} \rangle$ overwhelmingly often, even when the inputs do not contain people or places.

This failure mode, as well as recent research suggesting that increased training dataset diversity improves downstream generalization capabilities \cite{vinvl, clip, gpt3, pile}, prompted us to construct another large-scale pretraining dataset from various publicly available image-text datasets, including a heavily filtered subset of LAION \cite{laion}, Wikipedia Image-Text \cite{wit}, CC3M \cite{cc3m}, Visual Genome \cite{visualgenome}, Localized Narratives \cite{localizednarratives}.

Following research showing that LMs become strong zero-shot learners after being finetuned on collections of structured, task-based datasets \cite{flan, t0}, we also include the training splits of the following downstream tasks: VQA \cite{vqa}, GQA \cite{gqa}, OKVQA \cite{okvqa}, VizWiz \cite{vizwiz}, Hateful Memes \cite{hatefulmemes}, CoCo Captions \cite{cococaptions}.
This results in a dataset of around 25 million image-text pairs to train our final model, see §\ref{s:final_model}.


\section{Experiments and Analysis}\label{sec:expsandanalysis}

To evaluate our methodology, we first train a series of ablations (cf. §\ref{subsec:ablations}), to break down the effects of the vision encoder and adapter choice. We evaluate these ablations, and all subsequent models on a range of \textit{visual question answering} and \textit{image captioning} tasks designed to quantify the model's ability to adapt to new tasks using \textit{in-context} learning, recognize a wide variety of objects, and reason in detail about an image -- often involving complex spatial understanding, encyclopedic world knowledge, and optical character recognition (OCR).

\subsection{Evaluations} \label{subsec:evaluations}

\subsubsection{Visual Question Answering (VQA)} \label{subsec:vqa}
\begin{figure}[t]
    \centering
    \includegraphics[width=\linewidth]{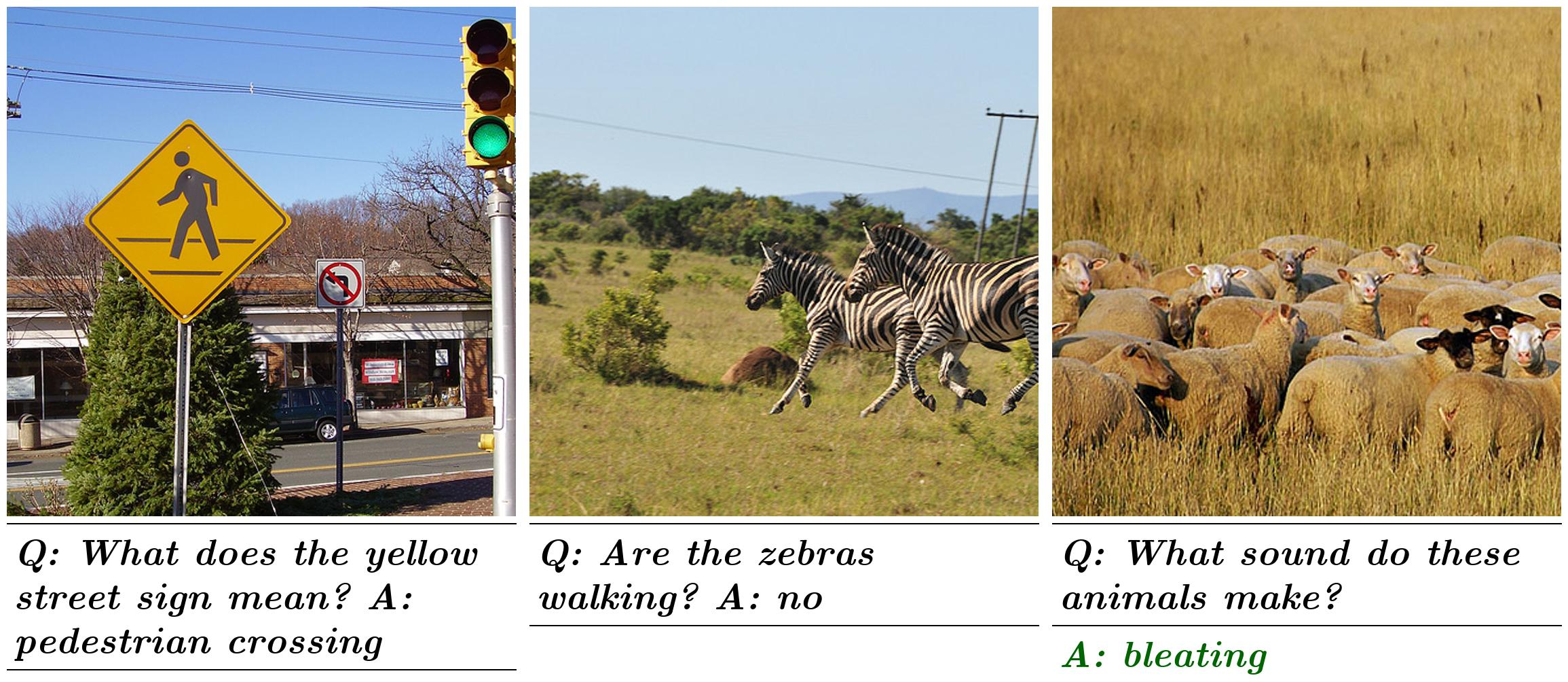}
    \caption{An example of a \textit{2-shot} prompt for OKVQA.}
    \label{fig:vqa}
\end{figure}

VQA tasks require the model to answer a question about the input image. Breaking from previous works, which generally formulate VQA tasks as classification tasks over the most frequent responses in the training set, we formulate all VQA tasks as open-ended generative tasks to enable few-shot prompting. We use the following datasets:

\noindent
\textbf{VQA 2.0}~\cite{vqa}. A large and commonly used dataset for VQA where samples consist of an image, a question regarding the content of the image and $10$ corresponding ground-truth answers.

\noindent
\textbf{OKVQA}~\cite{okvqa}. A VQA dataset where correct answers require explicit outside world knowledge not contained in the picture. 

\noindent
\textbf{GQA}~\cite{gqa}. A large VQA dataset focusing on visual and spatial reasoning.

\noindent
\textbf{VizWiz}~\cite{vizwiz}. A dataset in the same format as VQA with questions asked by visually impaired people. The ground-truth to a question about an image may be ``unanswerable'' or ``unsuitable'', which has to be recognized by the model.

To compare the generated model output with the provided ground-truths, we apply the normalization procedure of the official VQA 2.0 repo, \footnote{\url{https://github.com/GT-Vision-Lab/VQA}} and truncate the model output to the length of the longest ground truth answer. For VQA, OKVQA, and VizWiz
we calculate the accuracy metric from the official VQA paper \cite{vqa}, and for GQA we use the canonical accuracy score.

For few-shot settings, we use the procedure described in \citet{frozen}, prepending $n$ random examples of completed tasks before each question answer pair. We preprend "Q:~" and "A:~" to each question and answer respectively, improving performance (as exemplified in Figure \ref{fig:vqa}).

\subsubsection{Image Captioning} \label{subsec:captioning}

Image captioning tasks require the model to generate accurate descriptions of input images in natural language. We evaluate on two datasets -- CoCo Captions \cite{cococaptions} and NoCaps \cite{nocaps}, measuring performance using the BLEU@4 and CIDEr metrics. NoCaps is designed to evaluate a model's ability to caption images containing uncommon or novel object classes that don't appear in CoCo.

Like with \textit{SimVLM}, prompting with ``A picture of'' dramatically increases downstream scores, e.g. for $\mathrm{MAGMA}_{long}$, increasing the CIDEr score on CoCo Captions from $7.5$ to $57.1$. All scores reported in Table \ref{t:ablations} use this as a prefix. Other prefixes,
such as ``Caption:'' have a similar effect.

\subsubsection{Visual Entailment}
We test Visual Entailment performance on SNLI-VE \cite{snli_ve}, a task built on top of SNLI \cite{SNLI}. SNLI-VE requires the model to reason about the relationship between an \textit{image premise}, $P_{\mathrm{image}}$, and a \textit{text hypothesis}, $H_{\mathrm{text}}$. Given $P_{\mathrm{image}}$ and $H_{\mathrm{text}}$ as input, the task is to label their relationship as either entailment, neutral or contradiction. We formulate SNLI-VE as a classification task by finetuning the model together with a linear classification head
on the last-layer transformer embedding of the last text token.

\begin{figure}[t]
    \centering
    \includegraphics[width=0.85\linewidth]{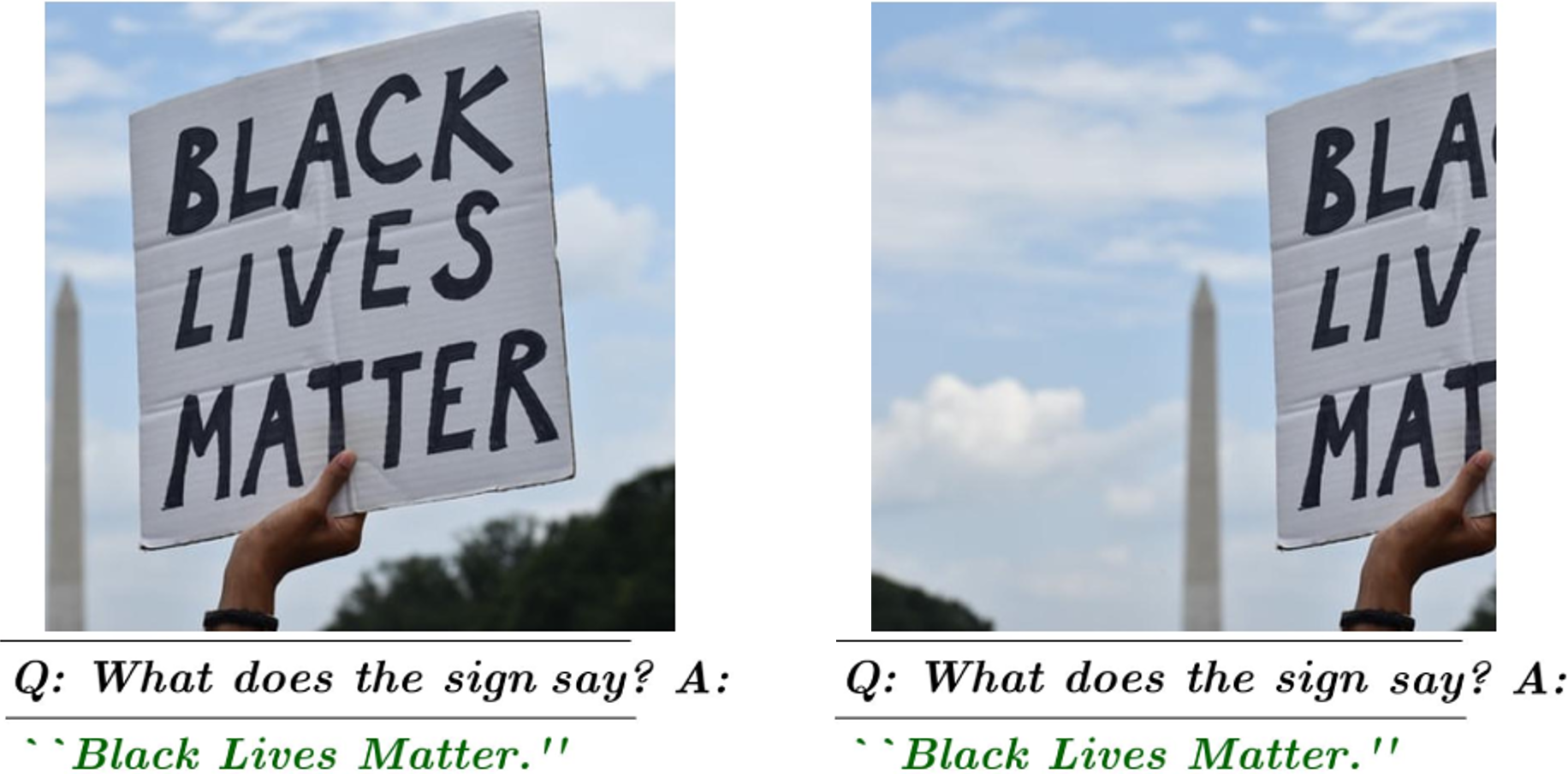}
    \caption{$\mathrm{MAGMA}$'s OCR capabilities. Even when text is obscured, $\mathrm{MAGMA}$ imputes the missing values.}
    \label{fig:blm}
\end{figure}

\subsection{Ablations} \label{subsec:ablations}

We run two series of ablations: i) One designed to test the impact of the adapter layers and their precise configuration, and ii) another designed to test the impact of the vision encoder choice. We also independently replicate the \textit{Frozen} model (see Table~\ref{t:ablations}), using the pretraining setup described in their paper (with the exception that we pretrain on CC12M) to use as a baseline. All ablations are trained for a total of 15k steps, or around 3.8 million image-text pairs.

\begin{table*}[t!] 
    \centering
    \resizebox{\linewidth}{!}{
 \begin{tabular}{|c c c c c|c c c c|c c c c|c c c c|c c c c|c|} 
 \hline
\multicolumn{5}{|c|}{\textbf{Adapter ablations}} & \multicolumn{4}{c|}{$n$-shot-VQA} & \multicolumn{4}{c|}{$n$-shot-OKVQA} & \multicolumn{4}{c|}{$n$-shot-GQA} & \multicolumn{4}{c|}{$n$-shot-VizWiz} & Avg. \\[0.5ex] 
 Type & $\lambda$ & Attn & FF & Params &  $0$ & $1$ & $2$ & $4$ & $0$ & $1$ & $2$ & $4$ & $0$ & $1$ & $2$ & $4$ & $0$ & $1$ & $2$ & $4$ & \\
 \hline\hline
-- & -- & -- & -- & $0.1$ &  $36.4$ & $41.5$ & $41.7$ & $41.8$ & $12.5$ & $16.2$ & $16.0$ & $16.5$ & $12.6$ & $\mathbf{20.8}$ & $23.6$ & $26.9$ & $2.9$ & $5.3$ & $5.5$ & $6.7$ & $20.4$ \\ 
 s & $1$ & -- & $2$ & $2$ & $34.7$ & $40.1$ & $42.2$ & $43.2$ & $12.4$ & $16.9$ & $18.6$ & $21.5$ & $8.2$ & $14.1$ & $19.2$ & $24.6$ & $5.3$ & $7.4$ & $7.8$ & $9.7$ & $20.4$\\ 
 s & $1$ & -- & $4$ & $1$ & $32.7$ & $40.2$ & $42.5$ & $43.8$ & $11.7$ & $16.3$ & $19.1$ & $21.2$ & $6.8$ & $15.6$ & $22.1$ & $27.7$ & $4.2$ & $6.7$ & $6.9$ & $8.6$ & $20.0$\\
 s & $1$ & $8$ & $8$ & $1$ & $36.6$ & $41.7$ & $\mathbf{43.8}$ & $45.2$ & $\mathbf{13.9}$ & $17.1$ & $20.0$ & $22.5$ & $\mathbf{14.3}$ & $20.7$ & $\mathbf{24.9}$ & $\mathbf{28.4}$ & $\textbf{5.6}$ & $8.5$ & $8.6$ & $9.8$ & $\mathbf{22.6}$\\
 s & $1$ & $12$ & $6$ & $1$ & $\mathbf{36.9}$ & $41.2$ & $43.6$ & $44.7$ & $13.9$ & $\mathbf{19.4}$ & $\mathbf{21.6}$ & $23.2$ & $12.8$ & $18.8$ & $22.5$ & $25.8$ & $5.3$ & $\mathbf{9.6}$ & $\textbf{9.8}$ & $\mathbf{10.6}$ & $22.5$ \\
 p & $1$ & -- & $4$ & $1$ & $36.5$ & $41.7$ & $43.1$ & $43.8$ & $14.5$ & $18.4$ & $20.3$ & $21.8$ & $11.2$ & $16.3$ & $19.9$ & $23.2$ & $4.6$ & $8.4$ & $8.4$ & $9.2$ & $21.3$\\
 p & t & $8$ & $8$ & $1$ & $34.9$ & $\mathbf{42.2}$ & $44.1$ & $\mathbf{45.4}$ & $12.9$ & $17.7$ & $21.4$ & $\mathbf{23.4}$ & $8.8$ & $15.6$ & $20.2$ & $24.5$ & $4.3$ & $7.9$ & $8.5$ & $9.9$ & $21.4$\\
 [0.5ex] 
 \hline\hline
 \multicolumn{5}{|c|}{\textbf{Encoder ablations}} & \multicolumn{4}{c|}{} & \multicolumn{4}{c|}{} & \multicolumn{4}{c|}{} & \multicolumn{4}{c|}{} &  \\[0.5ex]
 \hline
 \multicolumn{5}{|c|}{NFResnet}  &  $32.0$ & $37.0$ & $39.0$ & $39.7$ & $9.8$ & $15.8$ & $18.9$ & $20.8$ & $9.1$ & $\mathbf{20.2}$ & $\mathbf{27.1}$ & $\mathbf{28.7}$ & $2.8$ & $5.6$ & $6.5$ & $8.2$ & $20.1$ \\ 
  \multicolumn{5}{|c|}{CLIP-ViT}  & $32.8$ & $33.9$ & $36.7$ & $37.7$ & $10.5$ & $9.2$ & $12.4$ & $14.2$ & $8.4$ & $14.9$ & $22.2$ & $25.7$ & $2.7$ & $5.1$ & $5.2$ & $7.7$ & $17.5$ \\ 
  \multicolumn{5}{|c|}{CLIP-RN50x4} & $\mathbf{35.2}$ & $40.0$ & $\mathbf{42.6}$ & $\mathbf{44.2}$ & $\mathbf{12.6}$ & $\mathbf{17.7}$ & $19.0$ & $\mathbf{21.8}$ & $\mathbf{10.5}$ & $13.0$ & $16.1$ & $20.5$ & $\mathbf{5.0}$ & $6.2$ & $6.6$ & $8.3$ & $20.0$ \\
  \multicolumn{5}{|c|}{CLIP-RN50x16} & $32.7$ & $\mathbf{40.2}$ & $42.5$ & $43.8$ & $11.7$ & $16.3$ & $\mathbf{19.1}$ & $21.2$ & $6.8$ & $15.6$ & $22.1$ & $27.7$ & $4.2$ & $\mathbf{6.7}$ & $\mathbf{6.9}$ & $\mathbf{8.6}$ & $\mathbf{20.4}$ \\
  \hline\hline
  \multicolumn{5}{|c|}{Frozen (NFResnet + no adapters)} & $28.6$ & $36.7$ & $37.9$ & $38.1$ & $6.2$ & $15.1$ & $16.2$ & $15.8$ & $8.7$ & $23.5$ & $27.0$ & $27.5$ & $1.7$ & $5.4$ & $6.2$ & $8.0$ & $18.9$ \\
  \hline\hline
  \multicolumn{5}{|c|}{\textbf{MAGMA pretrained}} & \multicolumn{4}{c|}{} & \multicolumn{4}{c|}{} & \multicolumn{4}{c|}{} & \multicolumn{4}{c|}{} &  \\[0.5ex]
  \hline
  \multicolumn{5}{|c|}{$\mathrm{MAGMA}_{base}$} & $60.0$ & -- & -- & -- & $37.6$ & -- & -- & -- & $47.4$ & -- & -- & -- & $15.9$ & -- & -- & -- & $40.3$ \\
  \multicolumn{5}{|c|}{$\mathrm{MAGMA}_{long}$} & $\mathbf{61.5}$ & -- & -- & -- & $\mathbf{40.3}$ & -- & -- & -- & $\mathbf{49.6}$ & -- & -- & -- & $\mathbf{16.7}$ & -- & -- & -- & $\mathbf{42.0}$ \\
  \hline\hline\hline
  \multicolumn{5}{|c|}{\textbf{Adapter ablations}} & \multicolumn{4}{c|}{NoCaps - CIDEr} & \multicolumn{4}{c|}{NoCaps - B@4} & \multicolumn{4}{c|}{CoCo - CIDEr} & \multicolumn{4}{c|}{CoCo - B@4} &  \\[0.5ex] 
 Type & $\lambda$ & Attn & FF & params &  In & Out & Near & All & In & Out & Near & All & \multicolumn{4}{c|}{} & \multicolumn{4}{c|}{} & \\
 \hline\hline
 -- & -- & -- & -- & $0.1$ &  $45.1$ &  $53.7$ & $43.3$  & $45.7$ & $9.9$ & $5.8$ & $7.9$ &  $7.8$ &  \multicolumn{4}{c|}{$36.7$}  & \multicolumn{4}{c|}{$10.3$} & \\ 
   s & $1$ & -- & $2$ & $2$ &  $37.7$ & $55.5$ & $40.6$ &  $43.2$ & $6.2$ & $6.1$ & $6.5$ &  $6.4$ &  \multicolumn{4}{c|}{$33.4$}  & \multicolumn{4}{c|}{$9.4$} & \\
   s & $1$ & -- & $4$ & $1$ &  $39.3$ & $56.2$ & $44.0$ & $45.8$ & $6.3$ & $6.7$ & $7.7$ &  $7.3$ &  \multicolumn{4}{c|}{$39.6$}  & \multicolumn{4}{c|}{$11.2$} & \\
    s & $1$ & $8$ & $8$ & $1$ &  $38.2$ & $49.5$ & $40.9$ &  $42.2$ & $6.4$ & $4.9$ & $6.7$ &  $6.3$ &  \multicolumn{4}{c|}{$37.1$}  & \multicolumn{4}{c|}{$10.6$} & \\
     s & $1$ & $12$ & $6$ & $1$ &  $\mathbf{51.9}$ & $\mathbf{64.8}$ & $\mathbf{54.6}$ &  $\mathbf{56.2}$ & $\mathbf{11.4}$ & $\mathbf{8.4}$ & $\mathbf{11.3}$ &  $\mathbf{10.8}$ &  \multicolumn{4}{c|}{$\mathbf{46.3}$}  & \multicolumn{4}{c|}{$\mathbf{13.9}$} & \\
      p & $1$ & -- & $4$ & $1$ & $37.5$ & $38.1$ & $35.9$ &  $36.0$ & $7.2$ & $5.1$ & $6.7$ &  $6.4$ &  \multicolumn{4}{c|}{$36.3$}  & \multicolumn{4}{c|}{$10.8$} & \\
 p & t & $8$ & $8$ & $1$ &  $40.6$ & $58.3$ & $45.0$ &  $47.1$ & $8.0$ & $6.6$ & $7.9$ &  $7.7$ &  \multicolumn{4}{c|}{$39.5$}  & \multicolumn{4}{c|}{$11.2$} & \\ [0.5ex]
 \hline\hline
  \multicolumn{5}{|c|}{\textbf{Encoder ablations}} & \multicolumn{4}{c|}{} & \multicolumn{4}{c|}{} & \multicolumn{4}{c|}{} & \multicolumn{4}{c|}{} &  \\[0.5ex]
 \hline
 \multicolumn{5}{|c|}{NFResnet}  &  $22.5$ & $16.2$ & $22.0$ &  $20.9$ & $5.0$ & $1.6$ & $5.3$ &  $4.5$ &  \multicolumn{4}{c|}{$22.4$}  & \multicolumn{4}{c|}{$8.2$} & \\
  \multicolumn{5}{|c|}{CLIP-ViT}  & $33.2$ & $44.2$ & $35.3$ &  $36.8$ & $5.9$ & $5.2$ & $5.8$ &  $5.7$ &  \multicolumn{4}{c|}{$27.2$}  & \multicolumn{4}{c|}{$7.7$} & \\
  \multicolumn{5}{|c|}{CLIP-RN50x4} & $\mathbf{47.7}$ & $43.6$ & $\mathbf{48.1}$ &  $\mathbf{50.2}$ & $\mathbf{9.3}$ & $6.7$ & $\mathbf{9.2}$ &  $\mathbf{8.7}$ &  \multicolumn{4}{c|}{$\mathbf{41.9}$}  & \multicolumn{4}{c|}{$\mathbf{13.1}$} & \\
  \multicolumn{5}{|c|}{CLIP-RN50x16} &  $39.3$ & $\mathbf{56.2}$ & $44.0$ &  $45.8$ & $6.3$ & $\mathbf{6.7}$ & $7.7$ &  $7.3$ &  \multicolumn{4}{c|}{$39.6$}  & \multicolumn{4}{c|}{$11.2$} & \\ \hline\hline
  \multicolumn{5}{|c|}{\textbf{MAGMA pretrained}} & \multicolumn{4}{c|}{} & \multicolumn{4}{c|}{} & \multicolumn{4}{c|}{} & \multicolumn{4}{c|}{} &  \\[0.5ex]
  \hline
  \multicolumn{5}{|c|}{$\mathrm{MAGMA}_{base}$} & $55.8$ & $56.5$ & $49.9$ & $52.1$ & $11.1$ & $6.1$ & $10.3$ &  $9.5$ & \multicolumn{4}{c|}{$51.1$} & \multicolumn{4}{c|}{$15.8$} &  \\
  \multicolumn{5}{|c|}{$\mathrm{MAGMA}_{long}$} & $\mathbf{58.1}$& $\mathbf{62.0}$ & $\mathbf{56.9}$ &  $\mathbf{58.1}$ & $\mathbf{13.3}$ & $\mathbf{8.5}$ & $\mathbf{13.2}$ &  $\mathbf{12.3}$ & \multicolumn{4}{c|}{$\mathbf{57.0}$} & \multicolumn{4}{c|}{$\mathbf{17.6}$} &  \\ [0.5ex]
  \hline
 \end{tabular}
}
\caption{Performance evaluation on downstream tasks.
Open-ended few-shot evaluation on VQA-val, OKVQA-val, GQA-testdev and VizWiz-val. Captioning evaluation on NoCaps-val and CoCo-val. Models under \textbf{MAGMA pretrained} are trained on the mixed dataset detailed in Section~\ref{subsec:dataset}, all other models are trained on CC12M. \\
Notation for adapter ablations. \textbf{Type}: (s)caled or (p)arallel. $\bm{\lambda}$: $1$ or (t)rained. \textbf{Attn, FF}: Downsample factor of the bottleneck in the resp. position. -- means not applied. \textbf{Params}: Number of trainable parameters relative to the ablation with sequential FF adapters with downsample factor $4$.}
\label{t:ablations}
\end{table*}

\begin{table*}[h] 
\centering
\footnotesize

 \begin{tabular}{|c| c c c c |c|c c |c c|} 
 \hline
 & VQA & OKVQA & GQA & VizWiz & SNLI-VE & \multicolumn{2}{c|}{NoCaps} & \multicolumn{2}{c|}{Coco} \\
& & & & & & CIDEr & B@4 & CIDEr & B@4 \\[0.5ex] 
 \hline\hline
$\mathrm{MAGMA}$  & $68.0$ & $\mathbf{49.2}$ & $54.5$ & $35.4$ & $79.0$ & $93.6$ & $27.8$ & $91.2$ & $31.4$ \\
SOTA & $\mathbf{75.5}$ & $48.0$ & $\mathbf{72.1}$ & $\mathbf{54.7}$ & $\mathbf{86.3}$ & $\mathbf{112.2}$ &  	$\mathbf{33.1}$ & $\mathbf{143.3}$ & $\mathbf{41.7}$  \\
SOTA model & \textit{SimVLM} & \textit{PICa} & \textit{CFR} & \textit{Pythia} & \textit{SimVLM} & \textit{SimVLM} & \textit{VIVO} & \textit{SimVLM} & \textit{OSCAR} \\
 [0.5ex] 
 \hline
 \end{tabular}
 \caption{$\mathrm{MAGMA}$ finetuned performance.
\textbf{B@4}: NoCaps-all score. SOTA scores are to the best of our knowledge at the time of writing. If available/applicable, we compare to the SOTA score of models solving the task in an open-ended generative fashion like $\mathrm{MAGMA}$ (notably \textit{SimVLM} on VQA), otherwise we compare to the general SOTA (classification setting).
Models: \textit{SimVLM} \cite{simvlm}, \textit{PICa} \cite{gpt3_okvqa}, \textit{CFR} \cite{cfr}, \textit{Pythia} \cite{pythia}, \textit{VIVO} \cite{vivo}, \textit{OSCAR} \cite{oscar}. 
 } 
\label{t:finetuning}
\end{table*}


\subsubsection{Adapter Types}\label{subsubsec:adapter_ablations}

We run ablations with several different adapter configurations, motivated by \citet{unified_adapters} showing that the precise formulation of the adapter layer can have a large impact on the performance of a model on downstream tasks. Also, different adapter layers can perform better than others depending on the task.
Since an exhaustive sweep in the parameter space of adapters is very expensive, we decided on seven configurations, including models with no adapters, to get
a qualitative picture of the effect on downstream performance.  We use the same visual encoder (CLIP `RN50x16') for all adapter ablations and evaluate the open-ended few-shot scores on the VQA and Image Captioning tasks described in \ref{subsec:vqa} and \ref{subsec:captioning} respectively. The results are shown in Table~\ref{t:ablations}. Although there is no adapter configuration which clearly outperforms the rest, we observe three key points:


\textbf{Applying adapters to the attention layer is key.} Adapter configurations with no adapters on the attention layer underperform, particularly at few shot prompting.

\textbf{More adapter parameters to the feed forward layer increases performance on knowledge-based tasks.} The adapter variant with more parameters allocated to the feed forward adapter outperforms other variants on OKVQA and NoCaps tasks requiring outside knowledge and uncommon object classes recognition respectively. This supports preliminary research indicating that the feed-forward blocks are important in storing implicit knowledge in pretrained transformers \cite{knowledgeneurons}.

\textbf{Balancing attention and feed-forward parameter allocation aids scene understanding.} The adapter variant with equal number of parameters allocated to the  attention and the feed forward  adapters excels at the GQA benchmark, a QA benchmark built around scene graphs and designed to focus on skills such as spatial reasoning, comparisons, and object and attribute recognition.

    \label{fig:factored}%

\subsubsection{Visual Encoders}

We run ablations with four different image encoders: NFResnet, CLIP-ViT-B/32, CLIP-RN50x4 and CLIP-RN50x16. All visual encoder
ablations
are trained using the adapter configuration with sequential adapters on the feed-forward block and a downsample factor of $4$. The results are shown in Table \ref{t:ablations}. Our findings are the following:

\textbf{CLIP-RN50x16, on average, performs best at VQA tasks}. However, the difference between RN50x16 and RN50x4 is slight, with the smaller encoder performing better on VQA and OKVQA, while the larger encoder has a much higher GQA accuracy. We hypothesize that the increased resolution of the larger feature grid results in a more detailed scene understanding, while the smaller grid is better at condensing global visual information, which also shows in the Image Captioning scores, where CLIP-RN50x4 excels.

\textbf{CLIP-ViT has the worst average score across question answering tasks.} This reinforces the finding of \citet{clip_vl}, who find that the CLIP-ViT model struggles at tasks which require localization within an image.

Recall that the image prefix length varies between image encoders which may have a confounding effect on the results -- further study is needed to disentangle the effects of sequence length and the choice of the vision encoder.

\subsection{Final Model} \label{s:final_model}


\begin{figure}[t]
    \centering
    \includegraphics[width=\linewidth]{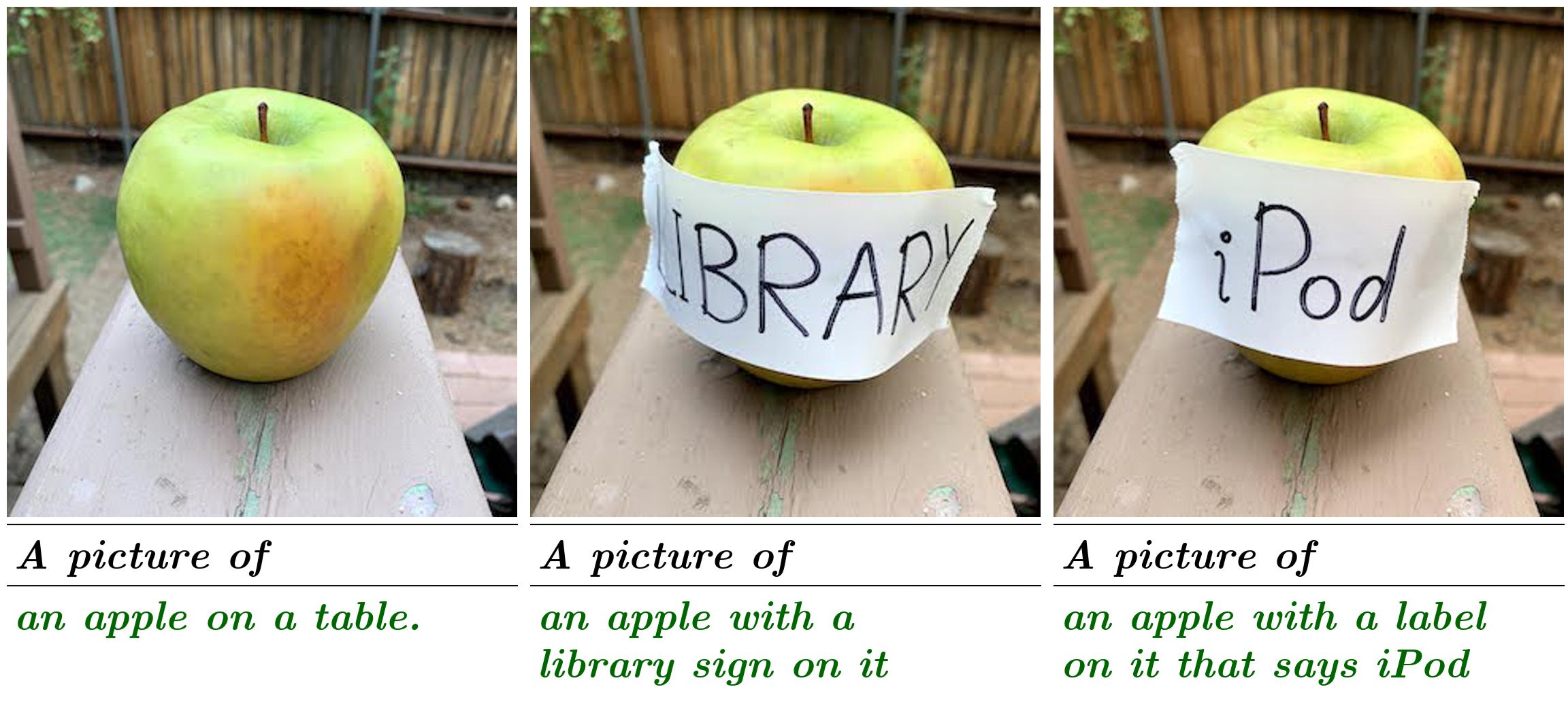}
    \caption{An example of an adversarial \textit{typographic attack} which MAGMA appears robust to, unlike CLIP.}
    \label{fig:apple}
\end{figure}

Based on our ablation studies, in particular the average VQA scores,  we opt to train a final $\mathrm{MAGMA}$ model 
using the CLIP-RN50x16 encoder and sequential adapters with a downsample factor of 8 applied to the feed-forward and attention layers. We train on the dataset detailed in §\ref{subsec:dataset} and see that evaluation loss does not plateau after $\sim$3M samples as reported in \textit{Frozen}, and so continue training, resulting in two model variants --  $\mathrm{MAGMA}_{base}$ trained for 15k steps for comparability to \textit{Frozen}, and $\mathrm{MAGMA}_{long}$ trained for $~$7.6M samples.

Due to the inclusion of training splits of tasks like VQA in the pretraining dataset, the performance of $\mathrm{MAGMA}_{base}$ significantly exceeds the downstream performance of previously trained ablations. The evaluation is conducted in the same way as the zero-shot procedure for the ablations and to avoid cluttered notation, we refer to it as such, although ``zero-shot`` usually refers to solving tasks unseen in pretraining. We stress that \emph{the pretraining set and the eval sets are still disjoint}.

While the scores of $\mathrm{MAGMA}_{long}$ already surpass the VQA-finetuned variants reported in \textit{Frozen}, we find that we can further increase the single-task performance on the training sets of each benchmark described in §\ref{subsec:evaluations} by finetuning on them. After finetuning, $\mathrm{MAGMA}$ achieves competitive scores across all benchmarks, setting a new state of the art accuracy on OKVQA, as well as attaining strong scores on the NoCaps benchmark -- to our knowledge, being surpassed only by \textit{SimVLM} and \textit{VinVL} \cite{vinvl}, see Table~\ref{t:finetuning}.

\begin{figure}[t]
    \centering
    \includegraphics[width=\linewidth]{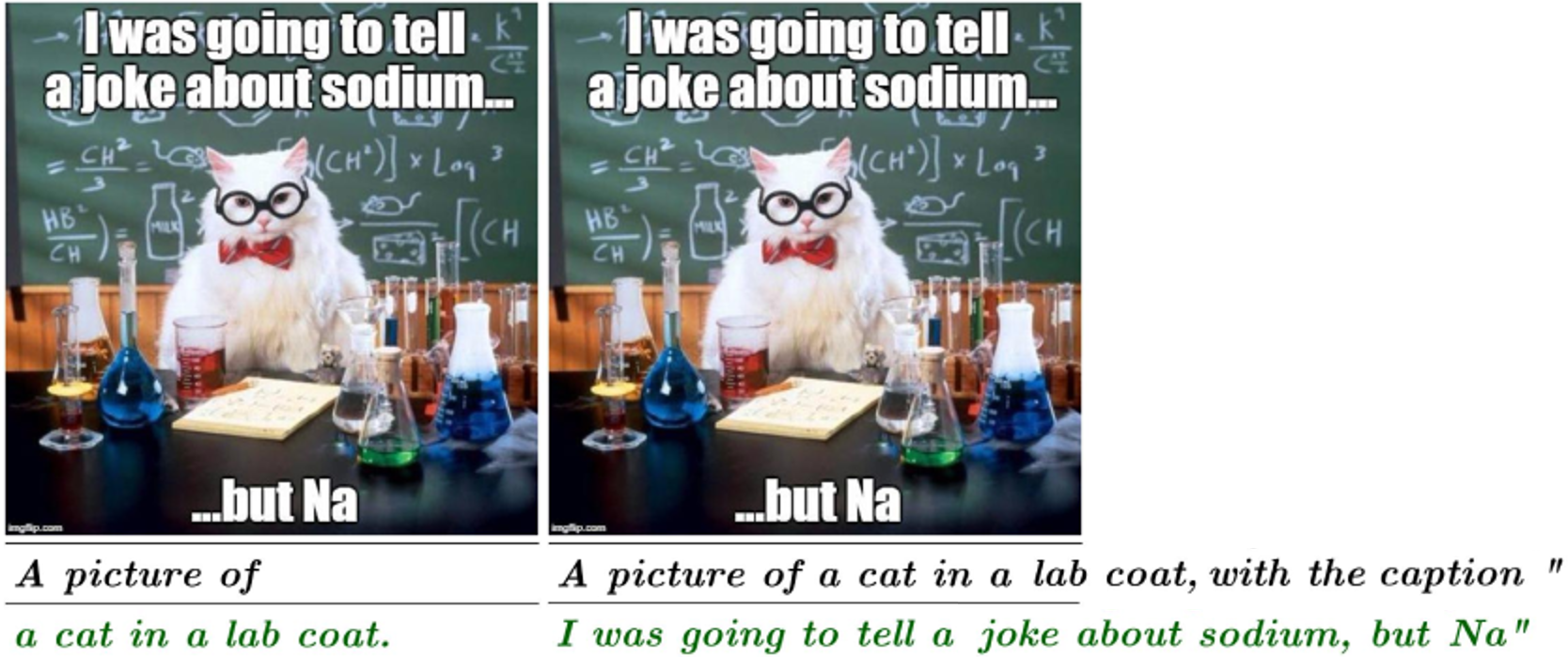}
    \caption{Example of \textit{multi-step prompting}. Using the output of the model (left) again as the input (right), the generation procedure is broken down into atomic steps.}
    \label{fig:my_label}
\end{figure}

We include several \textbf{qualitative results}, which highlight strengths of the model we feel are not sufficiently reflected by the evaluations in Table \ref{t:ablations}. Notably, $\mathrm{MAGMA}$ appears to be less easily fooled by the adversarial \textit{typographic attacks} to which CLIP is susceptible \cite{multimodalneurons}, see Figure \ref{fig:apple}. Additionally, $\mathrm{MAGMA}$ shows impressive OCR capabilities even without supervised finetuning, see Figure~\ref{fig:blm}, which warrants further quantitative evaluation. Interestingly, if a word or phrase is truncated, $\mathrm{MAGMA}$ can often impute the missing text. We also include an example of a multi-step \textit{factored cognition} prompt \cite{mishra2021reframing}, see Figure~\ref{fig:my_label}, where a challenging task is broken down into atomic steps. We suspect that task decomposition may enable $\mathrm{MAGMA}$ to perform complex tasks that it would otherwise be unable to solve.




\section{Conclusion}
We propose a simple framework for Multimodal Augmentation of Generative Models through Adapter-based Finetuning -- demonstrating that it is possible to transform multiple unimodal models into a powerful multimodal VL model while keeping the weights of the language component frozen. Our model, $\mathrm{MAGMA}$, trained using adapter layers and a simple next token prediction objective, can perform competitively with state of the art VL models on a wide range of benchmarks, excelling at tasks requiring external knowledge and the recognition of uncommon object classes. 

We hope our results will act as a starting point for further research into augmenting pretrained language models with additional modalities.

\section{Limitations}
Although the performance of $\mathrm{MAGMA}$ is impressive, we note some current limitations with the model and autoregressive VL models in general. Firstly, as we observed in the Image Captioning tasks, LMs can be sensitive to input -- performance is heavily dependent on the prompt format. 

Secondly, although the model can perform in-context learning with multiple examples in its context window, it struggles to reason over multiple images, as it was only pretrained on single image-caption pairs. 

Finally, $\mathrm{MAGMA}$ shows similar capabilities to large LMs like GPT3, about which there are ongoing ethical concerns regarding their reproduction of biases from the training data, as well as concerns relating to how to effectively align their outputs to human goals. As such, further research into the reproduction of visual biases, and the guiding of model outputs is needed.

\section*{Acknowledgements}
We would like to thank Mayukh Deb for his help with setting up and maintaining the public repository which makes MAGMA available to the research community. We would additionally like to thank the research team at Aleph Alpha for providing a stimulating and supportive environment.

\bibliography{anthology,custom,egbib}
\bibliographystyle{acl_natbib}

\clearpage

\appendix

\section{Appendix: Training details} \label{app:appendix}

During pretraining for the ablations and all subsequent models, we update the parameters $\theta$ by minimizing the loss~\eqref{e:loss_fn} per mini-batch using the Adam optimizer in combination with \textit{ZeRO} \cite{zero} to parallelize gradients and optimizer states across devices. We train all models with a batch size of 256, a dropout probability of 0.1, a weight decay of 0, and use learning rates of $2\cdot10^{-6}$ for $V^e_\theta$ and $8\cdot10^{-4}$ for $(V^p_\theta, \{A_{i,\theta} \})$, annealing both to 10\% of their original value using a cosine decay schedule throughout training. When finetuning on downstream tasks (see Section~\ref{s:final_model}) we do early stopping based on validation loss, and use the same hyperparameters as above, aside from decreasing the learning rates for $V^e_\theta$, $(V^p_\theta, \{A_{i,\theta} \})$ to $1.5 \cdot 10^{-6}$, $7 \cdot 10^{-4}$ for generative tasks and $1.5 \cdot 10^{-6}$, $3 \cdot 10^{-4}$ for SNLI-VE classification. We build our model using the PyTorch framework with Deepspeed for data-parallel training -- training all ablations on 32 A100 GPUs for around 1.25 days each.


\end{document}